\documentclass[twoside]{article}
\usepackage[accepted]{aistats2016}
\usepackage{times}
\usepackage{helvet}
\usepackage{courier}
\usepackage{wrapfig}

\usepackage{graphicx} 
\usepackage{subfigure} 
\usepackage{float}

\usepackage[numbers,square,sort&compress]{natbib}

\usepackage{url}

\usepackage{mathtools}
\usepackage{amsmath,amsfonts}

\usepackage[ruled,vlined]{algorithm2e}
\AtBeginDocument{
  \addtolength\abovedisplayskip{-0.1\baselineskip}
  \addtolength\belowdisplayskip{-0.15\baselineskip}
}

\DeclareMathOperator{\tr}{tr}
\DeclareMathOperator{\vect}{vec}
\DeclareMathOperator*{\argmin}{\arg\!\min}

\newcommand{\by}{\mathbf{y}}
\newcommand{\bx}{\mathbf{x}}
\newcommand{\bY}{\mathbf{Y}}
\newcommand{\bX}{\mathbf{X}}
\newcommand{\bLambda}{\mathbf{\Lambda}}
\newcommand{\bTheta}{\mathbf{\Theta}}
\newcommand{\bSig}{\mathbf{\Sigma}}
\newcommand{\bPsi}{\mathbf{\Psi}}
\newcommand{\bGamma}{\mathbf{\Gamma}}

\newcommand{\bSxx}{\mathbf{S_{xx}}}
\newcommand{\bSxy}{\mathbf{S_{xy}}}
\newcommand{\bSyy}{\mathbf{S_{yy}}}

\newcommand{\bDL}{\mathbf{D}_\bLambda}
\newcommand{\bDT}{\mathbf{D}_\bTheta}

\newcommand{\bU}{\mathbf{U}}
\newcommand{\bV}{\mathbf{V}}
\newcommand{\bR}{\mathbf{R}}

\begin{document}
\runningtitle{Large-Scale Optimization Algorithms for Sparse CGGMs}
\twocolumn[
\aistatstitle{Large-Scale Optimization Algorithms for \\ Sparse Conditional Gaussian Graphical Models}

\aistatsauthor{ Calvin McCarter \And Seyoung Kim }

\aistatsaddress{ Machine Learning Department \\ Carnegie Mellon University \And Computational Biology Department \\ Carnegie Mellon University } ]

\begin{abstract} 
This paper addresses the problem of scalable optimization for $l_1$-regularized 
conditional Gaussian graphical models.  Conditional Gaussian graphical models 
generalize the well-known Gaussian graphical models to conditional distributions 
to model the output network influenced by conditioning input variables. 
While highly scalable optimization methods exist for sparse 
Gaussian graphical model estimation, state-of-the-art methods for 
conditional Gaussian graphical models are not efficient enough 
and more importantly, fail due to memory constraints for very large problems. 
In this paper, we propose a new optimization procedure based on a Newton method 
that efficiently iterates over two sub-problems, leading to drastic improvement 
in computation time compared to the previous methods. 
We then extend our method to scale to large problems under memory constraints, 
using block coordinate descent to limit memory usage while achieving fast convergence. 
Using synthetic and genomic data, we show that our methods can solve problems with millions of variables and tens of billions of parameters to high accuracy on a single machine.
\end{abstract} 

\section{\uppercase{Introduction}}
\label{sec:introduction}

Sparse Gaussian graphical models (GGMs)~\citep{glasso} have been extremely popular
as a tool for learning a network structure over a large
number of continuous variables in many different application
domains including neuroscience~\citep{ng2011} and biology~\citep{glasso}. 
A sparse GGM can be estimated as a 
sparse inverse covariance matrix by
minimizing the convex function of $l_1$-regularized negative log-likelihood. 
Highly scalable learning algorithms such as graphical lasso~\citep{glasso}, 
QUIC~\citep{Hsieh:2011}, and BigQUIC~\citep{Hsieh:2013}
have been proposed to learn the model.


In this paper, we address the problem of scaling up the optimization of sparse conditional GGM (CGGM),
a model closely related to sparse GGM, to very large problem sizes without requiring excessive time or memory.
Sparse CGGMs have been introduced as a discriminative extension of sparse GGMs 
to model a sparse network over outputs conditional on input 
variables~\cite{Sohn:2012,Wytock:2013}. 
CGGMs can be viewed as a Gaussian analogue of conditional random field~\citep{crf}, 
while GGMs are a Gaussian analogue of Markov random field.
Sparse CGGMs have recently been applied to various settings 
including energy forecasting~\citep{Wytock:2013:energy}, finance~\citep{yuan2014}, and biology~\citep{Zhang:2014}, 
where the goal is to predict structured outputs influenced by inputs.
A sparse CGGM can be estimated by minimizing a convex
function of $l_1$-regularized negative log-likelihood.
This optimization problem is closely related to
that for sparse GGMs because CGGMs also model the network over outputs.
However, the presence of the additional parameters in CGGMs for the functional mapping
from inputs to outputs makes the optimization significantly more complex than in sparse GGMs.

Several different approaches have been previously proposed to estimate
sparse CGGMs, including OWL-QN~\citep{Sohn:2012}, 
accelerated proximal gradient method~\citep{yuan2014},
and Newton coordinate descent algorithm~\citep{Wytock:2013}. 
In particular, the Newton coordinate descent algorithm 
extends the QUIC algorithm~\cite{Hsieh:2011} for sparse GGM estimation to the case of CGGMs,
and has been shown to have superior computational speed and convergence.
This approach finds in each iteration a descent direction
by minimizing a quadratic approximation of the original negative
log-likelihood function along with $l_1$ regularization.
Then, the parameter estimate is updated
with this descent direction and a step size found by line search.

Although the Newton coordinate descent method~\cite{Wytock:2013} is state-of-the-art 
for its scalability and fast convergence, it is still not efficient enough 
to be applied to many real-world problems even with tens
of thousands of variables. More importantly,  
it suffers from a large space requirement, because 
for very high-dimensional problems,
several large dense matrices need to be precomputed and stored during optimization. 
For a CGGM with $p$ inputs and $q$ outputs,  
the algorithm requires storing several $p \times p$ and $q \times q$ dense matrices,
which cannot fit in memory for large $p$ and $q$.

We propose new algorithms for learning $l_1$-regularized CGGMs 
that significantly improve the computation time
of the previous Newton coordinate descent algorithm and also remove 
the large memory requirement.
We first propose an optimization method, 
called an alternating Newton coordinate descent algorithm, 
for improving computation time.
Our algorithm is based on the key observation that 
the computation simplifies drastically, if 
we alternately optimize the two sets of 
parameters for output network and for mapping inputs to outputs, 
instead of updating all parameters at once as in the previous approach.
The previous approach updated all parameters simultaneously 
by forming a second-order approximation of the objective on all parameters, which
requires an expensive computation of the large Hessian
matrix of size $(p+q)\times (p+q)$ in each iteration. 
Our approach of alternate optimization forms a second-order approximation
only on the network parameters, which requires the Hessian of size $q \times q$,
as the other set of parameters can be updated easily using a simple coordinate descent.

In order to overcome the constraint on the space requirement,
we then extend our algorithm to an alternating Newton block coordinate descent method
that can be applied to problems of unbounded size on a machine with limited memory.
Instead of recomputing each element of the large matrices on demand,
we divide the parameters into blocks for block-wise updates such that the results of computation 
can be reused within each block.
Block-wise parameter updates were previously used in BigQUIC~\citep{Hsieh:2013} 
for learning a sparse GGM, where the block sparsity pattern of 
the network parameters was leveraged to overcome the space limitations.
We propose an approach for block-wise update of the output network parameters in CGGMs that
extends their idea. We then propose a new block-wise update strategy
for the parameters for mapping inputs to outputs.
In our experiments, we show that we can solve problems with a million inputs and hundreds of
thousands of outputs on a single machine.

The rest of the paper is organized as follows. 
In Section \ref{sec:background}, we provide a brief review
of sparse CGGMs and  the current state-of-the-art Newton coordinate descent
algorithm~\cite{Wytock:2013} for learning the models.
In Section \ref{sec:alternating-cd}, we propose
an alternating Newton coordinate descent algorithm that 
significantly reduces computation time compared to the previous method.
In Section \ref{sec:alternating-bcd}, we further extend our algorithm
to perform block-wise updates in order to scale up to very large problems 
on a machine with bounded memory.
In Section \ref{sec:experiments}, we demonstrate our proposed algorithms 
on synthetic and real-world genomic data. 

\section{\uppercase{Background}}
\label{sec:background}


\subsection{The Conditional Gaussian Graphical Model}
A CGGM~\citep{Sohn:2012,Wytock:2013} models the conditional probability density of $\bx \in \mathbb{R}^{p}$ 
given $\by \in \mathbb{R}^{q}$ as follows:
\begin{align*}
p(\by|\bx; \bLambda,\bTheta) = \exp\{-\by^T\bLambda \by - 2\bx^T\bTheta \by\}/Z(\bx),
\end{align*} 
where $\bLambda$ is a $q \times q$ matrix for modeling the network
over $\by$ and $\bTheta$ is a $p \times q$ matrix for modeling the mapping
between the input variables $\bx$ and output variables $\by$. 
The normalization constant is given as 
$Z(\bx) = c |\bLambda|^{-1} \exp(\bx^T\bTheta\bLambda^{-1}\bTheta^T\bx)$.
Inference in a CGGM gives 
$
p(\by|\bx) = \mathcal{N}(\mathbf{B} \bx,\bLambda^{-1})
$, where $\mathbf{B}\!=\!-\bTheta\bLambda^{-1}$,
showing the connection between a CGGM and multivariate linear regression.

Given a dataset of $\bX \in \mathbb{R}^{n \times p}$
and $\bY \in \mathbb{R}^{n \times q}$ for $n$ samples,
and their covariance matrices
$\bSxx = \frac{1}{n}\bX^T \bX, \bSxy = \frac{1}{n}\bX^T \bY, \bSyy = \frac{1}{n}\bY^T \bY$,
a sparse estimate of CGGM parameters can be obtained 
by minimizing $l_1$-regularized negative log-likelihood:
\begin{align}
\min_{\bLambda \succ 0,\bTheta} & f(\bLambda,\bTheta) 
= g(\bLambda,\bTheta) + h(\bLambda,\bTheta), \label{eq:opt}
\end{align}
where 
$g(\bLambda, \bTheta) \!\!=\!\! -\!\log|\bLambda| \!+\!
\tr(\bSyy\bLambda \!+\! 2\bSxy^T\bTheta \!+\! 
\bLambda^{-1}\bTheta^T \bSxx \bTheta)$ for the smooth negative log-likelihood
and 
$h(\bLambda,\bTheta) \!\!=\!\! \lambda_{\bLambda} \|\bLambda\|_1 
+ \lambda_{\bTheta} \|\bTheta\|_1$
for the non-smooth elementwise $l_1$ penalty. 
$\lambda_\bLambda, \lambda_\bTheta > 0$ are regularization parameters. 
As observed in previous work~\citep{Sohn:2012,yuan2014,Wytock:2013}, this objective is convex.

\subsection{Optimization} 
The current state-of-the-art method for solving Eq. (\ref{eq:opt}) 
for $l_1$-regularized CGGM 
is the Newton coordinate descent algorithm~\citep{Wytock:2013} 
that extends QUIC~\citep{Hsieh:2011} for $l_1$-regularized GGM estimation. 
In each iteration, this algorithm found a generalized Newton descent direction
by forming a second-order approximation of the smooth part of the objective
and minimizing this along with the $l_1$ penalty.
Given this Newton direction, the parameter estimates were updated
with a step size found by line search using Armijo's rule~\citep{Armijo:1966}.

In each iteration, the Newton coordinate descent algorithm
found the Newton direction as follows:
\begin{align}
\bDL, \bDT = \argmin_{\Delta_\bLambda,\Delta_\bTheta} & \ 
\bar{g}_{\bLambda,\bTheta}(\Delta_\bLambda,\Delta_\bTheta) \nonumber \\
& + h(\bLambda\!+\!\Delta_\bLambda,\bTheta\!+\!\Delta_\bTheta),
\label{eq:nd}
\end{align}
where $\bar{g}_{\bLambda,\bTheta}$ is the second-order approximation of 
$g$ given by Taylor expansion:
\begin{eqnarray*}
\bar{g}_{\bLambda,\bTheta}(\Delta_\bLambda,\Delta_\bTheta) 
= \vect(\nabla g(\bLambda,\bTheta))^T \vect([\Delta_\bLambda \ \Delta_\bTheta])
\nonumber \\
\quad\quad +\frac{1}{2}\vect([\Delta_\bLambda \ \Delta_\bTheta])^T 
\nabla^2 g(\bLambda,\bTheta)
\vect([\Delta_\bLambda \ \Delta_\bTheta]). 
\end{eqnarray*}
The gradient and Hessian matrices above are given as:
\begin{eqnarray} 
\nabla g(\bLambda,\bTheta) =&
[\nabla_\bLambda g(\bLambda,\bTheta)\quad\quad  \nabla_\bTheta g(\bLambda,\bTheta) ]
\nonumber \\
=& [\bSyy \!-\! \bSig \!-\! \bPsi \quad 2\bSxy + 2\bGamma]
\label{eq:grad} \\
\nabla^2 g(\bLambda,\bTheta) =&
\begin{bmatrix}
\nabla^2_\bLambda g(\bLambda,\bTheta) & \nabla_\bLambda \nabla_\bTheta g(\bLambda,\bTheta) \\
\nabla_\bLambda \nabla_\bTheta g(\bLambda,\bTheta)^T & \nabla^2_\bTheta g(\bLambda,\bTheta) 
\end{bmatrix}
\nonumber \\
=& \begin{bmatrix}
\bSig \otimes (\bSig + 2 \bPsi)
& -2\bSig \otimes \bGamma^T \\
-2\bSig \otimes \bGamma
& 2\bSig \otimes \bSxx
\end{bmatrix}, \label{eq:hess}
\end{eqnarray}
where $\bSig = \bLambda^{-1}$, $\bPsi = \bSig\bTheta^T \bSxx \bTheta \bSig$,
 and $\bGamma = \bSxx\bTheta\bSig$.
Given the Newton direction in Eq. (\ref{eq:nd}), the parameters can be updated as 
$\bLambda \gets \bLambda + \alpha \bDL$ and $\bTheta \gets \bTheta + \alpha \bDT$,
where step size $0 < \alpha \le 1$ ensures
sufficient decrease in Eq. (\ref{eq:opt}) and positive definiteness of $\bLambda$.

The \textit{Lasso} problem~\citep{lasso} in Eq. (\ref{eq:nd})  
was solved using coordinate descent. 
Despite the efficiency of coordinate descent for \textit{Lasso}, 
applying coordinate updates repeatedly to all $q^2+pq$ variables in
$\bLambda$ and $\bTheta$ is costly. 
So, the updates were restricted to an active set of variables given as:
\begin{align}
\mathcal{S}_\bLambda \ &= \quad \{(\Delta_\bLambda)_{ij} : 
|(\nabla_\bLambda g(\bLambda,\bTheta))_{ij}| > \lambda_\bLambda \vee \bLambda_{ij} \ne 0\} 
\nonumber \\
\mathcal{S}_\bTheta \ &= \quad \{(\Delta_\bTheta)_{ij} : 
|(\nabla_\bTheta g(\bLambda,\bTheta))_{ij}| > \lambda_\bTheta \vee \bTheta_{ij} \ne 0\}.
\nonumber
\end{align}
Because the active set sizes 
$m_\bLambda = |\mathcal{S}_\bLambda|, m_\bTheta = |\mathcal{S}_\bTheta|$
approach the number of non-zero entries in the sparse solutions for
$\bLambda^*$ and $\bTheta^*$ over iterations,
this strategy yields a substantial speedup.

To further improve the efficiency of coordinate descent,
intermediate results were stored for  
the large matrix products that need to be computed repeatedly.
At the beginning of the optimization for Eq. (\ref{eq:nd}), 
$\bU := \Delta_\bLambda \Sigma$ and $\bV := \Delta_\bTheta \Sigma$
were computed and stored.
Then, after a coordinate descent update to $(\Delta_\bLambda)_{ij}$, 
row $i$ and $j$ of $\bU$ were updated. Similarly,
after an update to $(\Delta_\bTheta)_{ij}$, row $i$ of $\bV$ was updated.

\subsection{Computational Complexity and Scalability}

Although the Newton coordinate descent method is computationally more efficient than 
other previous approaches, it does not scale to problems even with tens of thousands of
variables. The main computational cost of the algorithm comes from computing
the large $(p+q) \times (p+q)$ Hessian matrix in Eq. (\ref{eq:hess}) 
in each application of Eq. (\ref{eq:nd}) to find the Newton direction.
At the beginning of the optimization in Eq. (\ref{eq:nd}),
large dense matrices $\bSig$, $\bPsi$, and $\bGamma$, for computing
the gradient and Hessian in Eqs. (\ref{eq:grad}) and (\ref{eq:hess}), are precomputed 
and reused throughout the coordinate descent iterations. 
Initializing $\bSig=\bLambda^{-1}$ via Cholesky decomposition costs up to $O(q^3)$ time,
although in practice, sparse Cholesky decomposition
exploits sparsity to invert $\bLambda$ in much less than $O(q^3)$.
Computing $\bPsi=\bR^T \bR$, where $\bR=\bX\bTheta\bSig$, 
requires $O(nm_\bTheta + nq^2)$ time, 
and computing $\bGamma$ costs $O(npq + nq^2)$.
After the initializations, the cost of coordinate descent update
per each active variable $(\Delta_\bLambda)_{ij}$ and $(\Delta_\bTheta)_{ij}$ is $O(p+q)$.
%
During the coordinate descent for solving Eq. (\ref{eq:nd}),
the entire $(p+q) \times (p+q)$ Hessian matrix in Eq. (\ref{eq:hess}) 
needs to be evaluated, whereas for the gradient in Eq. (\ref{eq:grad}) 
only those entries corresponding to the parameters in active sets
are evaluated.


A more serious obstacle to scaling up to problems with large $p$ and $q$
is the space required to store dense matrices $\bSig$ (size $q\!\times\!q$), 
$\bPsi$ (size $q\!\times\!q$), and $\bGamma$ (size $p\!\times\!q$). 
In our experiments on a machine with 104 Gb RAM, 
the Newton coordinate descent method exhausted memory when $p+q$ exceeded $80{,}000$.

In the next section, we propose a modification 
of the Newton coordinate descent algorithm
that significantly improves the computation time. 
Then, we introduce block-wise update strategies to our algorithm
to remove the memory constraint and to scale to arbitrarily large problem sizes.

\section{\uppercase{Alternating Newton Coordinate Descent}}
\label{sec:alternating-cd}

In this section, we introduce our alternating Newton coordinate descent algorithm 
for learning an $l_1$-regularized CGGM
that significantly reduces computation time compared to the previous method.
Instead of performing Newton descent 
for all parameters $\bLambda$ and $\bTheta$ simultaneously, 
our approach alternately updates $\bLambda$ and $\bTheta$, 
optimizing Eq. (\ref{eq:opt}) over $\bLambda$ given $\bTheta$ 
and vice versa until convergence.

Our approach is based on the key observation that with $\bLambda$ fixed,
the problem of solving Eq. (\ref{eq:opt}) over $\bTheta$ becomes
simply minimizing a quadratic function with $l_1$ regularization.
Thus, it can be solved efficiently using a coordinate descent method,
without the need to form a second-order approximation or to perform line search.
On the other hand, optimizing Eq. (\ref{eq:opt}) for $\bLambda$ given $\bTheta$ still requires forming a quadratic
approximation to find a generalized Newton direction and performing line search
to find the step size. However, this computation involves only $q \times q$ Hessian matrix and
is significantly simpler than
performing the same type of computation on both $\bLambda$ and $\bTheta$ jointly
as in the previous approach.


\begin{algorithm}[tb]
\caption{Alternating Newton Coordinate Descent}
\label{alg:Fast-alg}
   \DontPrintSemicolon
   \SetKwInOut{Input}{input}\SetKwInOut{Output}{output}
   \Input{Inputs $X \in \mathbb{R}^{n \times p}$ and $Y \in \mathbb{R}^{n \times q}$;
regularization parameters $\lambda_{\bLambda}, \lambda_{\bTheta}$}
   \Output{Parameters $\bLambda, \bTheta$}
   Initialize $\bTheta \leftarrow 0, \bLambda \leftarrow I_q$\;
   \For{$t=0,1,\dots$}{
     Determine active sets $\mathcal{S}_\bLambda, \mathcal{S}_\bTheta$\;
     Solve via coordinate descent:\;
     \quad $D_\bLambda \!=\! 
     \underset{\Delta_\bLambda,\Delta_{\overline{\mathcal{S}_\bLambda}=0}}{\argmin} 
     \bar{g}_{\bLambda,\bTheta}(\bLambda+\Delta_\bLambda,\!\bTheta) 
     + h(\bLambda\!+\!\Delta_\bLambda,\!\bTheta)$\;
     Update $\bLambda^+ = \bLambda + \alpha D_\bLambda$, 
     where step size $\alpha$ is found\; 
     \quad with line search\;
     Solve via coordinate descent:\;
     \quad $\bTheta^+ = \arg\!\min_{\bTheta_{\mathcal{S}_\bTheta}} \ g_\bLambda(\bTheta) + \lambda_{\bTheta} \|\bTheta\|_1$\; 
   }
\end{algorithm}

\subsection{Coordinate Descent Optimization for $\bLambda$}
Given fixed $\bTheta$,
the problem of minimizing the objective in Eq. (\ref{eq:opt})
with respect to $\bLambda$ becomes
\begin{eqnarray}
\argmin_{\bLambda \succ 0} g_\bTheta(\bLambda)
+ \lambda_\bLambda \|\bLambda\|_{1},
\label{eq:opt_lambda}
\end{eqnarray}
where $g_\bTheta(\bLambda)= -\log|\bLambda| + \tr(\bSyy\bLambda 
+ \bLambda^{-1} \bTheta^T \bSxx \bTheta)$.
In order to solve this, we first find a generalized Newton direction 
that minimizes the $l_1$-regularized quadratic
approximation of $g_\bTheta(\bLambda)$:
\begin{eqnarray}
\bDL = \argmin_{\Delta_\bLambda} 
\bar{g}_{\bLambda,\bTheta}(\Delta_\bLambda) 
+ \lambda_\bLambda \|\bLambda+\Delta_\bLambda\|_{1}, 
\label{eq:ancd_lambda}
\end{eqnarray}
where $\bar{g}_{\bLambda,\bTheta}(\Delta_\bLambda)$ is obtained from
a second-order Taylor expansion and is given as
\begin{eqnarray*}
\bar{g}_{\bLambda,\bTheta}(\Delta_\bLambda) 
= \vect(\nabla_\bLambda g(\bLambda, \bTheta))^T \vect(\Delta_\bLambda) 
\nonumber \\
+ \frac{1}{2}\vect(\Delta_\bLambda)^T 
\nabla^2_\bLambda g(\bLambda,\bTheta) \vect(\Delta_\bLambda). 
\end{eqnarray*}
The $\nabla_\bLambda g(\bLambda, \bTheta)$ and
$\nabla^2_\bLambda g(\bLambda,\bTheta)$ above are components of the gradient
and Hessian matrices corresponding to $\bLambda$ in Eqs. (\ref{eq:grad}) and (\ref{eq:hess}).
We solve the \textit{Lasso} problem in Eq. (\ref{eq:ancd_lambda}) via coordinate descent. 
Similar to the Newton coordinate descent method, 
we maintain $\bU := \Delta_\bLambda \bSig$ to reuse intermediate results 
of the large matrix-matrix product.
Given the Newton direction for $\bLambda$, 
we update $\bLambda \leftarrow \bLambda + \alpha \Delta_\bLambda$,
where $\alpha$ is obtained by line search. 

Restricting the generalized Newton descent to $\bLambda$
simplifies the computation significantly for coordinate descent updates, 
compared to the previous approach \citep{Wytock:2013}
that applies it to both $\bLambda$ and $\bTheta$ jointly. 
Our updates only involve $\nabla_\bLambda g(\bLambda,\bTheta)$ and
$\nabla_\bLambda^2 g(\bLambda,\bTheta)$, and
no longer involve $\nabla_\bTheta g(\bLambda,\bTheta)$ and
$\nabla_\bLambda\nabla_\bTheta g(\bLambda,\bTheta)$, eliminating
the need to compute the large $p \times q$ dense matrix $\bGamma$ in $O(npq + nq^2)$ time.
Our approach also reduces the computational cost for the coordinate descent update 
of each element of $\Delta_\bLambda$ from $O(p+q)$ to $O(q)$.

\subsection{Coordinate Descent Optimization for $\bTheta$}
With $\bLambda$ fixed, the optimization problem in Eq. (\ref{eq:opt}) with 
respect to $\bTheta$ becomes
\begin{eqnarray}
\argmin_\bTheta \  g_\bLambda(\bTheta) + \lambda_{\bTheta} \|\bTheta\|_1, 
\label{eq:ancd_theta}
\end{eqnarray}
where $g_\bLambda(\bTheta) = \tr(2\bSxy^T\bTheta + \bLambda^{-1}\bTheta^T \bSxx \bTheta)$. 
Since $g_\bLambda(\bTheta)$ is a quadratic function itself,
there is no need to form its second-order Taylor expansion
or to determine a step size via line search.
Instead, we solve 
Eq. (\ref{eq:ancd_theta}) directly with coordinate descent method,
storing and maintaining $\bV := \bTheta \bSig$.
Our approach reduces the computation time for updating $\bTheta$
compared to the corresponding computation in the previous algorithm.
We avoid computing the large $p\!\times \! q$ matrix $\bGamma$,
which had dominated overall computation time with $O(npq)$. 
Our approach also eliminates the need for line search for updating $\bTheta$.
Finally, it reduces the cost 
for each coordinate descent update in $\bTheta$ to 
$O(p)$, compared to $O(p + q)$ for the corresponding computation 
for $\Delta_{\bTheta}$ in the previous method.


Our approach is summarized in Algorithm \ref{alg:Fast-alg}. 
We provide the details of the coordinate descent update equations in Appendix.
For approximately solving Eqs. (\ref{eq:ancd_lambda}) and (\ref{eq:ancd_theta}),
we warm-start $\bLambda$ and $\bTheta$ from the results of the previous iteration
and make a single pass over the active set,
which ensures decrease in the objective in Eq. (\ref{eq:opt})
and reduces the overall computation time in practice.

\section{\uppercase{Alternating Newton Block Coordinate Descent}}
\label{sec:alternating-bcd}

The alternating Newton coordinate descent algorithm in the previous section
improves the computation time of the previous state-of-the-art method, but
is still limited by the space required to store large matrices during coordinate
descent computation. Solving Eq. (\ref{eq:ancd_lambda}) for updating $\bLambda$ 
requires precomputing and storing $q \times q$ matrices, $\bSig$ and $\bPsi$,
whereas solving Eq. (\ref{eq:ancd_theta}) for updating $\bTheta$ requires 
$\bSig$ and a $p \times p$ matrix $\bSxx$.
A naive approach to reduce the memory footprint would be to recompute 
portions of these matrices on demand
for each coordinate update, which would be very expensive. 

In this section we describe how our algorithm in the previous section
can be combined with block coordinate descent to scale up the optimization
to very large problems on a machine with limited memory. 
During coordinate descent optimization, we update blocks of 
$\bLambda$ and $\bTheta$ 
so that within each block, the computation of the large matrices
can be cached and re-used, where
these blocks are determined automatically by exploiting the sparse stucture.
For $\bLambda$, 
we extend the block coordinate descent approach in BigQUIC \citep{Hsieh:2013}
developed for GGMs to take into account the conditioning variables 
in CGGMs.
For $\bTheta$, we describe a new approach for block coordinate descent
update. 
Our algorithm can, in principle, be applied to problems of any size on a
machine with limited memory.

\begin{algorithm}[tb]
\caption{Alternating Newton Block Coordinate Descent}
\label{alg:Huge-alg}
   \DontPrintSemicolon
   \SetKwInOut{Input}{input}\SetKwInOut{Output}{output}
   \SetKwComment{Comment}{$\triangleright$\ }{}
   \Input{Inputs $X \in \mathbb{R}^{n \times p}$ 
and outputs $Y \in \mathbb{R}^{n \times q}$;
regularization parameters $\lambda_{\bLambda}, \lambda_{\bTheta}$}
   \Output{Parameters $\bLambda, \bTheta$}
   Initialize $\bTheta \leftarrow 0, \bLambda \leftarrow I_q$\;
   \For{$t=0,1,\dots$}{
     Determine active sets $\mathcal{S}_\bLambda, \mathcal{S}_\bTheta$\;
     Partition columns of $\bLambda$ into $k_\bLambda$ blocks\;
     \Comment*[c]{Minimize over $\bLambda$}
     Initialize $\Delta_\bLambda \leftarrow 0$\;
     \For{$z=1$ \KwTo $k_\bLambda$}{
       Compute $\bSig_{C_z}, \bU_{C_z}$, and $\bPsi_{C_z}$\;
       \For{$r=1$ \KwTo $k_\bLambda$}{
         \If{$z \ne r$}{
           Identify columns $B_{zr} \subset C_r$ with active\;
           \quad elements in $\bLambda$\;
           Compute $\bSig_{B_{zr}}, \bU_{B_{zr}}$, and $\bPsi_{B_{zr}}$\;
         }
         Update all active $(\Delta_\bLambda)_{ij}$ in $(C_z,C_r)$\;
       }
     }
     Update $\bLambda^+ \leftarrow \bLambda + \alpha\Delta_\bLambda$,
     	where step size $\alpha$ is found\;
        \quad with line search.\;
     Partition columns of $\bTheta$ into $k_\bTheta$ blocks\;
     \Comment*[c]{Minimize over $\bTheta$}
     \For{$r=1$ \KwTo $k_\bTheta$}{
       Compute $\bSig_{C_r}$, and initialize $\bV \leftarrow \bTheta\bSig_{C_r}$\;
       \For{{\upshape row} $i \in \{1,\dots,p\}$ {\upshape if} $I_\phi(\mathcal{S}_{(i,C_r)})$}{
         Compute $(\bSxx)_{ij}$ for non-empty rows $j$ in $V_{C_r}$\;
         Update all active $\bTheta_{ij}$ in $(i,C_r)$\;
       }
     }
   }
\end{algorithm}

\begin{figure}[t]
\centering
\raisebox{-0.5\height}{\includegraphics[scale=1.2]{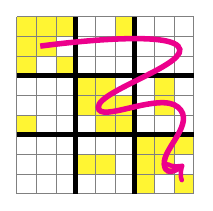}}
\ \ \
\raisebox{-0.5\height}{\includegraphics[scale=0.7]{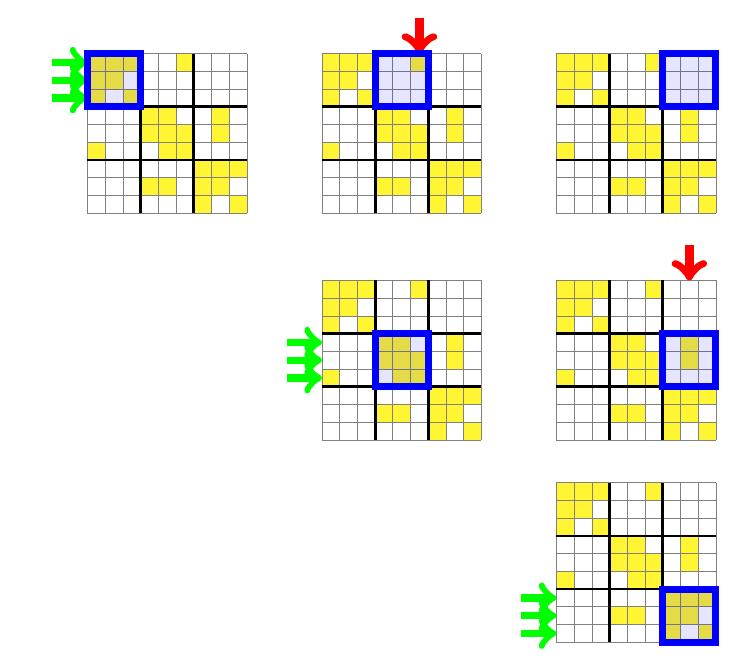}}
\vspace{-10pt}
\caption{
Schematic of block coordinate descent for $\bLambda$. 
The $\bLambda$ of size $q=9$ is updated for each of the $k_\bLambda^2$ blocks in turn with $k_\bLambda=3$.
Filled elements denote the parameters in the active set. 
The green arrows denote rows of $\bSig$ and $\bPsi$ that are computed once and reused
while sweeping through a row of blocks. The red arrows denote cache misses and the corresponding columns of
$\bSig$ and $\bPsi$ need to be recomputed.
}
\vspace{-5pt}
\label{fig:lambda-schema-grid}
\end{figure}

\begin{figure}[t]
\centering
\begin{tabular}{c@{\hspace{25pt}}c@{\hspace{10pt}}c}
\raisebox{-0.5\height}{\includegraphics[scale=1.5]{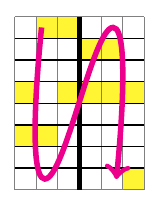}}
&
\raisebox{-0.5\height}{\includegraphics[scale=0.65]{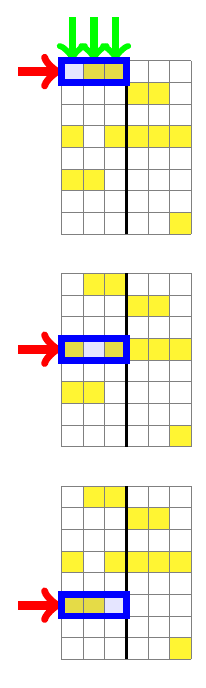}}
&
\raisebox{-0.5\height}{\includegraphics[scale=0.65]{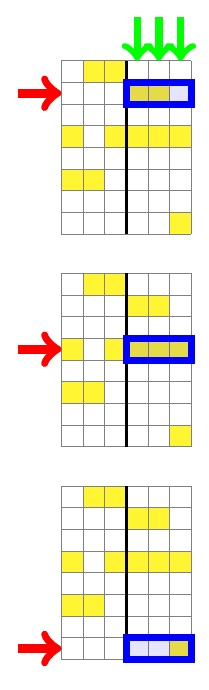}}
\end{tabular}
\vspace{-5pt}
\caption{
Schematic of block coordinate descent for $\bTheta$.  The $\bTheta$ of size $p\!=\!8,q\!=\!6$
is updated for each of the $p \times k_\bTheta$ blocks with $k_\bTheta\!=\!2$. 
Filled elements denote the parameters in the active set. 
Green arrows denote columns of $\bSig$ that are computed once and reused while sweeping 
through the column of $p$ blocks. The red arrows denote cache misses for $(\bSxx)_i$.
%
}
\vspace{-5pt}
\label{fig:theta-schema-grid}
\end{figure}

\subsection{Blockwise optimization for $\bLambda$}

\subsubsection{Block coordinate descent method}

A coordinate-descent update of $(\Delta_\bLambda)_{ij}$ 
requires the $i$th and $j$th columns of $\bSig$ and $\bPsi$.
If these columns are in memory, they can be
reused. Otherwise, it is a cache miss and we should compute them on demand.
$\bSig_i$ for the $i$th column of $\bSig$ can be obtained
by solving linear system $\bLambda \bSig_i = \mathbf{e}_i$ with
conjugate gradient method in $O(m_\bLambda K)$ time, 
where $K$ is the number of conjugate gradient iterations. 
Then, $\bPsi_i$ can be obtained from 
$\mathbf{R}^T \mathbf{R}_i$, where $\mathbf{R} = \bX \bTheta \bSig$
in $O(nq)$ time.


In order to reduce cache misses, we perform block coordinate descent,
where within each block, the columns of $\bSig$ are cached and re-used.
Suppose we partition $\mathcal{N} = \{1,\dots,q\}$ into $k_\bLambda$ blocks, 
$C_1,\dots,C_{k_\bLambda}$. We apply this partitioning to the rows and columns of 
$\Delta_\bLambda$ to obtain $k_\bLambda \times k_\bLambda$ blocks. 
We perform coordinate-descent updates in each block,
updating all elements in the active set within that block. 
Let $\mathbf{A}_{C_r}$ denote a $q$ by $|C_r|$ matrix 
containing columns of $\mathbf{A}$ that corresponds to the subset $C_r$. 
In order to perform coordinate-descent updates on $(C_r, C_z)$ block of $\Delta_\bLambda$,
we need $\bSig_{C_r}$, $\bSig_{C_z}$, $\bPsi_{C_r}$, and $\bPsi_{C_r}$. 
%
Thus, we pick the smallest possible $k_\bLambda$ such that we can store
$2q/k_\bLambda$ columns of $\bSig$ 
and $2q/k_\bLambda$ columns of $\bPsi$ 
in memory. 
%
When updating the variables within block $(C_z, C_r)$ of $\Delta_\bLambda$, 
there are no cache misses 
once $\bSig_{C_z}$, $\bSig_{C_z}$, $\bPsi_{C_z}$, and $\bPsi_{C_r}$
are computed and stored. 
After updating each $(\Delta_\bLambda)_{ij}$ to $(\Delta_\bLambda)_{ij} + \mu$, 
we maintain ${\bU}_{C_z}$ and $\bU_{C_r}$ by
$\bU_{it} \leftarrow \bU_{it} + \mu \bSig_{jt}, \bU_{jt} \leftarrow 
\bU_{jt} + \mu\bSig_{it}, \forall t \in \{C_z \cup C_r\}$.

To go through all blocks, we update blocks $(C_z, C_1), \dots, (C_z, C_k)$ 
for each $z \in \{1,\dots,k_\bLambda\}$. 
Since all of these blocks share $\bSig_{C_z}$ and $\bPsi_{C_z}$,
we precompute and store them in memory. 
When updating an off-diagonal block $(C_z, C_r), z \neq r$, 
we compute $\bSig_{C_r}$ and $\bPsi_{C_r}$.
In the worst case, overall
$\bSig$ and $\bPsi$ will be computed $k_\bLambda$ times. 

\subsubsection{Reducing computational cost using graph clustering}
In typical real-world problems, the graph structure of $\bLambda$ 
will exhibit clustering, with an approximately block diagonal structure.
We exploit this structure by choosing a partition 
$\{C_1,\dots, C_{k_\bLambda}\}$ that reduces cache misses.
Within diagonal blocks $(C_z, C_z)$'s, once $\bSig_{C_z}$ and $\bPsi_{C_z}$ 
are computed, there are no cache misses. 
For off-diagonal blocks $(C_z, C_r)$'s, $r \neq z$, 
we have a cache miss only if some variable in 
$\{\Delta_{ij} | i \in C_z, j \in C_r \}$ lies in the active set. 
We thus minimize the active set in off-diagonal blocks 
via clustering, following the strategy for sparse GGM estimation
in \citep{Hsieh:2013}.
In the best case, if all parameters in the active set appear in the diagonal blocks, 
$\bSig$ and $\bPsi$ are computed only once with no cache misses.
We use the METIS \citep{Karypis:1995} graph clustering library.
Our method for updating $\bLambda$ is illustrated in Figure \ref{fig:lambda-schema-grid}.

\subsection{Blockwise Optimization for $\bTheta$}
\label{subsec:bcd-bTheta}

\subsubsection{Block coordinate descent method}

The coordinate descent update of $\bTheta_{ij}$ 
requires $(\bSxx)_i$ and $\bSig_{j}$ to compute $(\bSxx)_i^T \bV_j$, 
where $\bV_{j} = \bTheta \bSig_{j}$.
If $(\bSxx)_i$ and $\bSig_{j}$ are not already in the memory, it is a cache miss. 
Computing $(\bSxx)_i$ takes $O(np)$, which is expensive 
if we have many cache misses. 

We propose a block coordinate descent approach for solving Eq. (\ref{eq:ancd_theta})
that groups these computations to reduce cache misses.
Given a partition $\{1,\dots,q\}$ into 
$k_\bTheta$ subsets, $C_1,\dots,C_{k_\bTheta}$, 
we divide $\bTheta$ into $p \times k_\bTheta$ blocks, 
where each block comprises a portion of a row of $\bTheta$.
We denote each block $(i, C_r)$, where $i \in \{1,\dots,p\}$.
Since updating block $(i,C_r)$ requires $(\bSxx)_i$ and $\bSig_{C_r}$, 
we pick smallest possible $k_\bTheta$ such that we can store 
$q/k_\bTheta$ columns of $\bSig$ in memory. 
%
While performing coordinate descent updates within block $(i,C_r)$ of $\bTheta$, 
there are no cache misses, once $(\bSxx)_i$ and $\bSig_{C_r}$ are in memory. 
After updating each $\bTheta_{ij}$ to $\bTheta_{ij}+\mu$, 
we update $\bV_{C_r}$ by 
$\bV_{it} \leftarrow \bV_{it} + \mu \bSig_{jt}, \forall t \in C_r$.

In order to sweep through all blocks,
each time we select a $q \in \{1,\dots,k_\bTheta\}$ and update blocks $(1,C_r), \dots, (p, C_r)$. 
Since all of these $p$ blocks with the same $C_r$ share the computation of $\bSig_{C_r}$, 
we compute and store $\bSig_{C_r}$ in memory. 
Within each block, the computation of $(\bSxx)_i$ is shared, so we pre-compute and store
it in memory, before updating this block.
The full matrix of $\bSig$ will be computed once while sweeping through the full $\bTheta$,
whereas in the worst case $\bSxx$ will be computed $k_\bTheta$ times. 

\subsubsection{Reducing computational cost using row-wise sparsity}

We further reduce cache misses for $(\bSxx)_i$
by strategically selecting partition $C_1,\dots,C_{k_\bTheta}$,
based on the observation that
if the active set is empty in block $(i,C_r)$,
we can skip this block and forgo computing $(\bSxx)_i$.
We therefore choose a partition
where the active set variables are clustered into as few blocks as possible.
Formally, we want to minimize $\sum_{i,q}{|I_\phi(\mathcal{S}_{(i,C_r)})|}$,
where $I_\phi(\mathcal{S}_{(i,C_r)})$ is an indicator function that outputs
1 if the active set $\mathcal{S}_{(i,C_r)}$ within block $(i,C_r)$
is not empty.
We therefore perform graph clustering over
the graph $G=(V,E)$ defined from the active set in $\bTheta$,
where $V = \{1, \ldots, q\}$ with one node for each column of $\bTheta$,
and $E = \{(j,k)|
\bTheta_{ij} \in \mathcal{S}_{\bTheta},
\bTheta_{ik} \in \mathcal{S}_\bTheta
\textrm{ for } i = 1, \ldots, p\}$,
connecting two nodes $j$ and $k$ with an edge
if both $\bTheta_{ij}$ and $\bTheta_{ik}$ are in the active set.
This edge set corresponds to the non-zero elements of $\bTheta^T \bTheta$,
so the graph can be computed quickly in $O(m_\bTheta q)$.

We also exploit row-wise sparsity in $\bTheta$ to
reduce the cost of each cache miss.
Every empty row in $\bTheta$ corresponds to an empty row in $\bV = \bTheta \bSig$.
Because we only need elements in $(\bSxx)_i$ for the dot product $(\bSxx)^T_i \bV_j$,
we skip computing the $k$th element of $(\bSxx)_{i}$ 
if the $k$th row of $\bTheta$ is all zeros.
Our strategy for updating $\bTheta$ is illustrated in Figure \ref{fig:theta-schema-grid}.


Our method is summarized in Algorithm \ref{alg:Huge-alg}. 
See Appendix for analysis of the computational cost.

\subsection{Parallelization}

The most expensive computations in our algorithm are embarrassingly parallelizable,
allowing for further speedups on machines with multiple cores. 
Throughout the algorithm, we parallelize matrix and vector multiplications.
In addition, for block-wise $\bLambda$ updates, we 
compute multiple columns of $\bSig_{C_z}$ and $\bPsi_{C_z}$ 
as well as multiple columns of $\bSig_{C_r}$ and $\bPsi_{C_r}$ for multiple cache misses in parallel,
running multiple conjugate gradient methods in parallel. 
For block-wise $\bTheta$ updates, we compute
multiple columns of $\bSig$ in parallel before sweeping through blocks
and perform a parallel compuation within each cache miss, computing elements within each $\bSxx_i$ in parallel.


\section{\uppercase{Experiments}}
\label{sec:experiments}
We compare the performance of our proposed methods with the existing state-of-the-art
Newton coordinate descent algorithm, using synthetic and real-world genomic
datasets. All methods were implemented in C++ with parameters represented in sparse matrix format.
All experiments were run on 2.6GHz Intel Xeon E5 machines with 
8 cores and 104 Gb RAM, running Linux OS.
We run the Newton coordinate descent and alternating Newton coordinate descent
algorithms as a single thread job on a single core.
For our alternating Newton block coordinate descent method, 
we run it on a single core and with parallelization on 8 cores.

\subsection{Synthetic Experiments}

\begin{figure*}
\centering
\begin{tabular}{@{}c@{}c@{}c}
\includegraphics[scale=0.47]{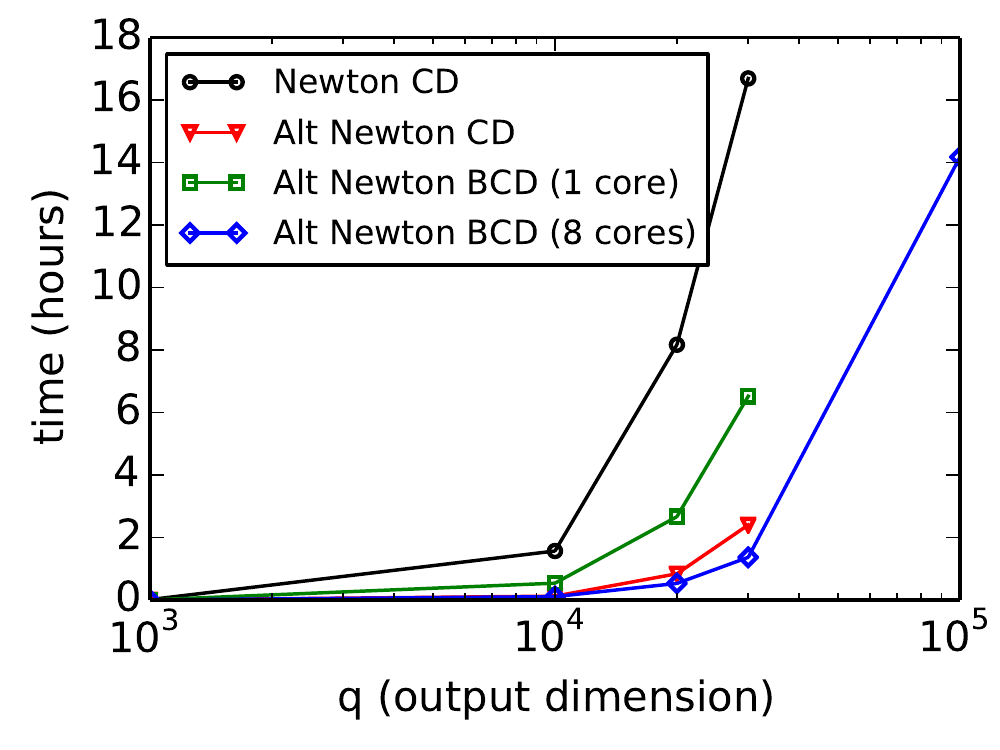}
& \includegraphics[scale=0.47]{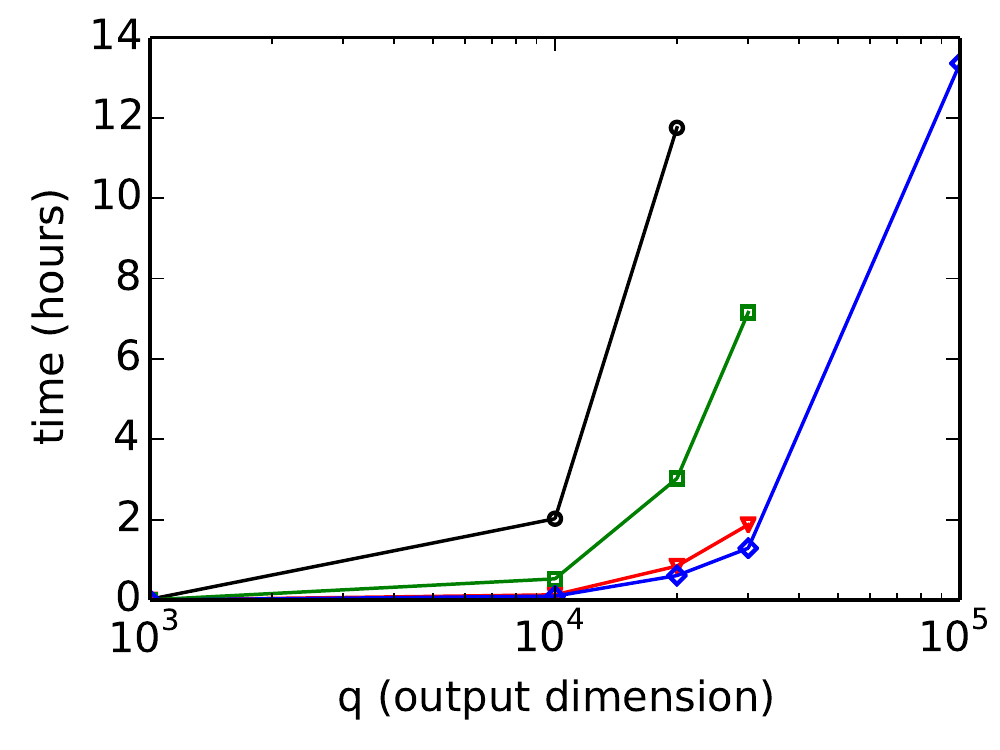}
& \includegraphics[scale=0.47]{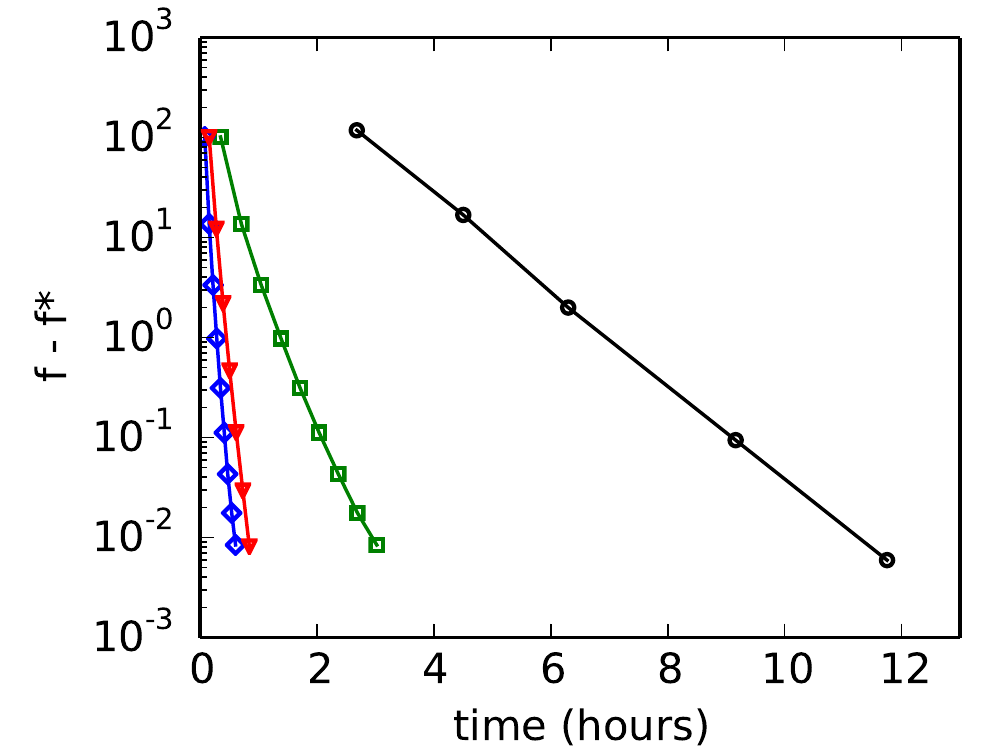} 
\vspace{-3pt}
\\
\vspace{-5pt}
(a) & (b) & (c)
\end{tabular}
\caption{Comparison of scalability on chain graphs.
We vary $p$ and $q$, where (a) $p=q$ and (b) $p=2q$.
The Newton coordinate descent and alternating Newton coordinate
descent methods could not be run beyond the problem sizes
shown due to memory constraint.
(c) Convergence when $q=20{,}000$ and $p=40{,}000$.
}
\label{fig:chain-results}
\vspace{-4pt}
\end{figure*}
\begin{figure*}
\centering
\begin{tabular}{@{}c@{}c@{}c}
\includegraphics[scale=0.47]{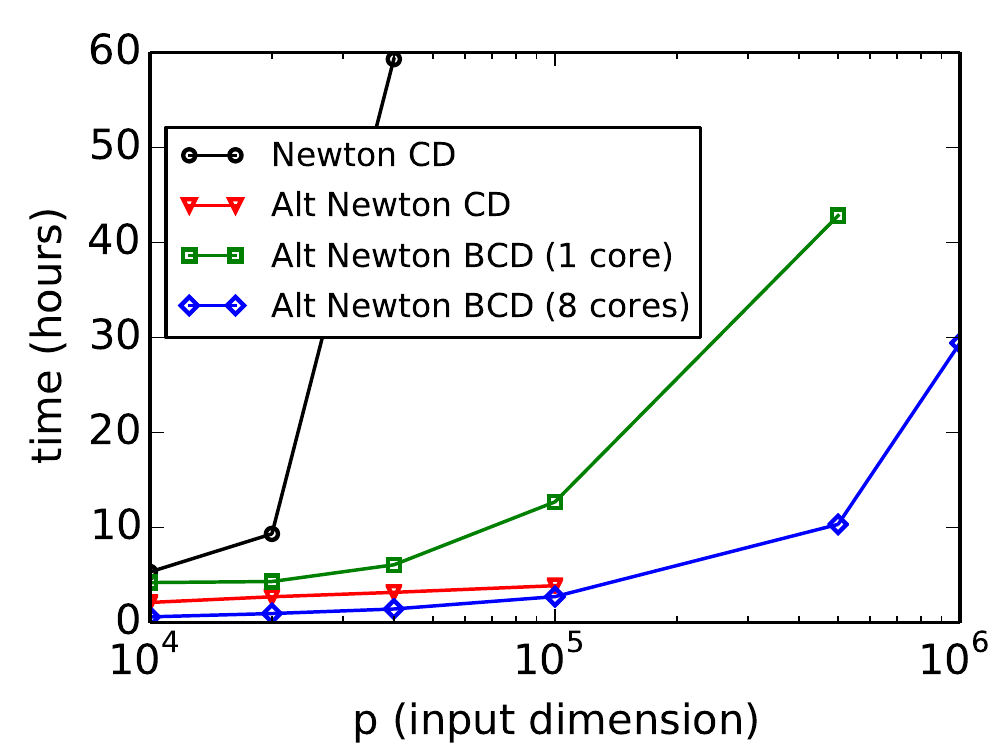}
& \includegraphics[scale=0.47]{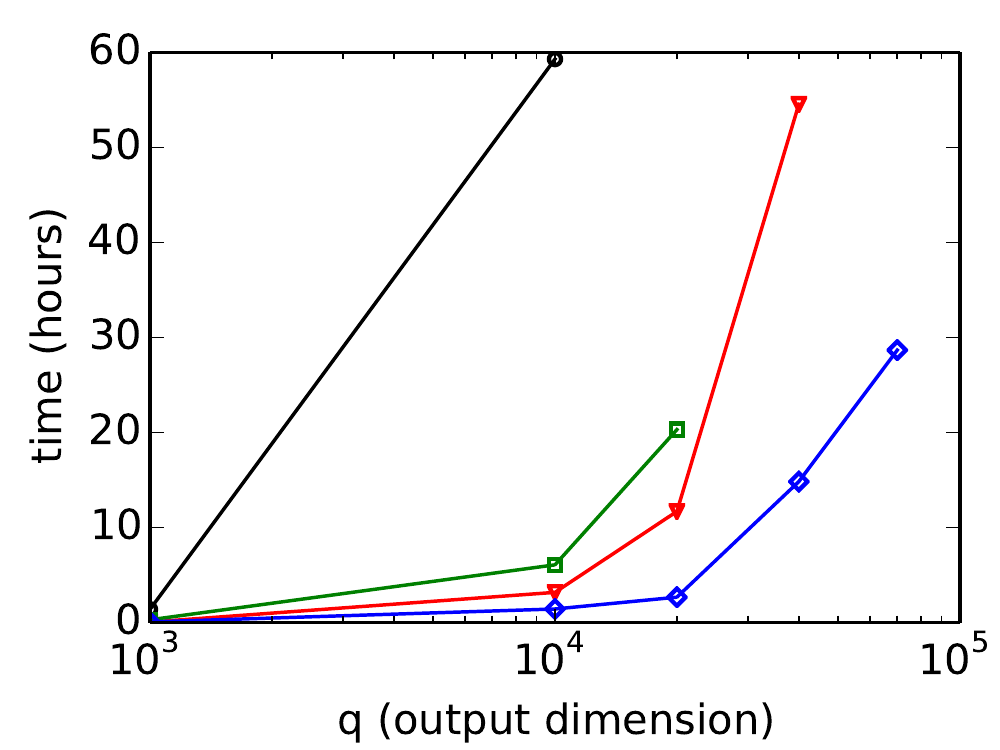}
& \includegraphics[scale=0.47]{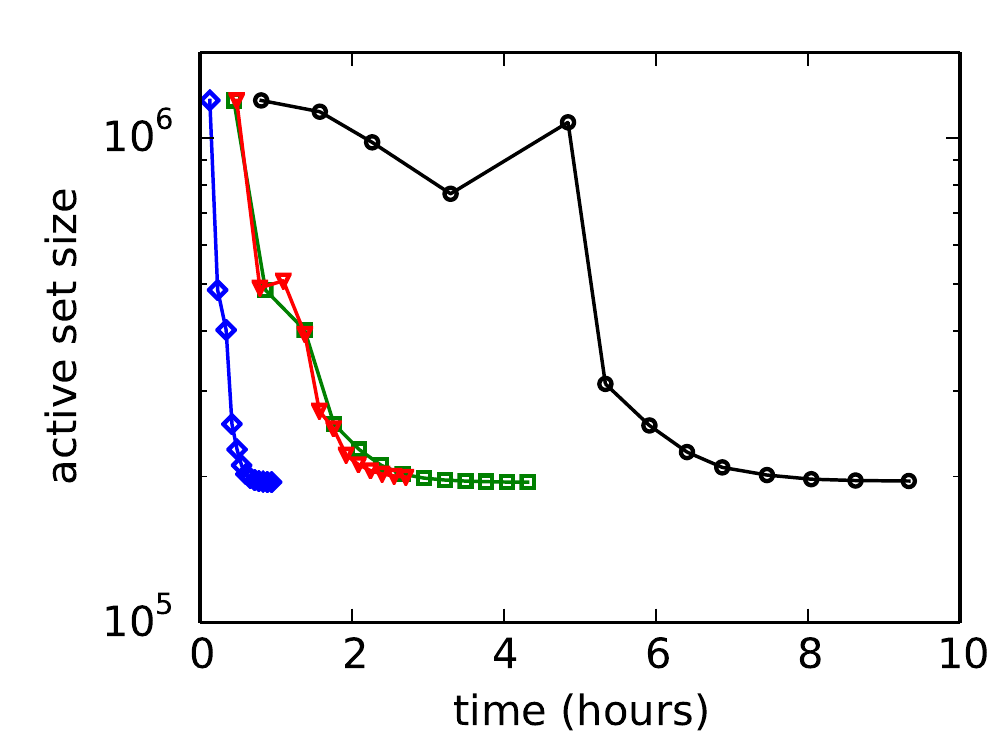} 
\vspace{-4pt}
\\
\vspace{-4pt}
(a) & (b) & (c)
\end{tabular}
\vspace{-4pt}
\caption{Comparison of scalability on random graphs with clustering.
(a) Varying $p$ with $q$ fixed at $10{,}000$.
(b) Varying $q$ with $p$ fixed at $40{,}000$.
(c) Active set size versus time with $p=20{,}000$ and $q=10{,}000$. 
}
\label{fig:cluster-results}
\vspace{-4pt}
\end{figure*}

We compare the different methods on two sets
of synthetic datasets, one for chain graphs 
and another for random graphs with clustering for $\bLambda$, generated as follows.
For chain graphs, the true sparse parameters $\bLambda$ is set 
with $\bLambda_{i,i-1}=1$ and $\bLambda_{i,i}=2.25$ and 
the ground truth $\bTheta$ is set with $\bTheta_{i,i}=1$.
We perform one set of chain graph experiments where the number of inputs $p$
equals the number of outputs $q$, 
and another set of experiments with an additional $q$ irrelevant features
unconnected to any outputs, so that $p$ equals $2q$.
For random graphs with clustering,
following the procedure in \citep{Hsieh:2013} for generating a GGM,
we set the true $\bLambda$ to a graph with
clusters of nodes of size $250$ and with $90\%$ of edges connecting
randomly-selected nodes within clusters. 
We set the number of edges so that 
the average degree of each node is 10, with edge weights set to 1. 
We then set the diagonal values
so that $\bLambda$ is positive definite.
To set the sparse patterns for $\bTheta$, we randomly select
$100\sqrt{p}$ input variables as having edges to at least one output
and distribute total $10q$ edges among those selected inputs to influence
randomly selected outputs.
We set the edge weights of $\bTheta$ to 1. 

Then, we draw samples from the CGGM defined by these true $\bLambda$ and $\bTheta$.
We generate datasets with $n=100$ samples for the chain graphs and
$n=200$ samples for random graphs with clustering.
We choose $\lambda_\bLambda$ and $\lambda_\bTheta$ so that the number of 
estimated edges in $\bLambda$ and $\bTheta$ is close to ground truth.
Following the strategy used in GGM estimation \citep{Hsieh:2013}, 
we use the minimum-norm subgradient of the objective as our stopping criterion:
$\|\text{grad}^S(\bLambda_t,\bTheta_t)\|_1 < 0.01(\|\bLambda\|_1 + \|\bTheta\|_1)$.

We compare the scalability of the different methods for chain graph experiments,
as we vary the problem sizes.
We show the computation times for datasets with $p=q$
in Figure \ref{fig:chain-results}(a) and for datasets with $p=2q$
with $q$ additional irrelevant features in Figure \ref{fig:chain-results}(b). 
For large problems, computation times are not shown for 
Newton coordinate descent and alternating Newton coordinate descent methods
because they could not complete the optimization with limited memory.
Also, for large problems, alternating block coordinate descent was terminated 
after 60 hours of computation.
We provide more results on varying the sample size $n$ in Appendix.
In Figure \ref{fig:chain-results}(c), using the dataset with $p=40{,}000$ and $q=20{,}000$, 
we plot the suboptimality in the objective $f - f^*$, 
where $f^*$ is obtained by running alternating Newton
coordinate descent algorithm to numerical precision.
Our new methods converge substantially faster than the previous approach,
regardless of desired accuracy level.
We notice that as expected from the convexity of the optimization problem, 
all algorithms converge to the global optimum and find nearly identical parameter estimates.

%

\begin{table*}[t!]
\caption{
Computation time in hours on genomic dataset.
`*' indicates running out of memory.
}
\centering
\label{table:real}
\begin{tabular}{@{}c@{\hskip 0.15in}c@{\hskip 0.15in}c@{\hskip 0.15in}c@{\hskip 0.15in}c@{\hskip 0.15in}c@{\hskip 0.15in}c}
\hline
$p$ & $q$ & $\|\bLambda^*\|_0$ & $\|\bTheta^*\|_0$ & Newton CD & Alternating Newton CD & Alternating Newton BCD \\
\hline
34{,}249 & 3{,}268 & 34{,}914 & 28{,}848 & 22.0 & 0.51 & \textbf{0.24} \\
34{,}249 & 10{,}256 & 86{,}090 & 103{,}767 & $>50$ & 2.4 & \textbf{2.3} \\
442{,}440 & 3{,}268 & 26,232 & 30,482 & * & * & \textbf{11} \\
\hline
\vspace{-15pt}
\end{tabular}
\end{table*}

In Figure \ref{fig:cluster-results}, we compare scalability 
of different methods for random graphs with clustering.
In Figure \ref{fig:cluster-results}(a), we vary $p$, while setting $q$ to $10{,}000$.
In Figure \ref{fig:cluster-results}(b),
we vary $q$, fixing $p$ to $40{,}000$.
Similar to the results from chain graph,
for larger problems,
Newton coordinate descent and alternating Newton coordinate descent methods
ran out of memory and alternating block coordinate descent was terminated 
after 60 hours.
For all problem sizes, our alternating Newton coordinate descent algorithm
significantly reduces the computation time of the previous method,
the Newton coordinate descent algorithm. This gap in the computation time
increases for larger problems.
In Figure \ref{fig:cluster-results}(c),
we compare the convergence in sparsity pattern for the different methods
as measured by the active set size,
for problem size $p=20{,}000$ and $q=10{,}000$.
All our methods recover the optimal sparsity pattern 
much more rapidly than the previous approach.

Figures \ref{fig:chain-results} and \ref{fig:cluster-results} 
show that our alternating Newton block coordinate descent
can run on much larger problems than any other methods, while those methods without block 
coordinate descent run out of memory. 
For example, in Figure \ref{fig:cluster-results}(a) alternating Newton block coordinate 
descent could handle problems with one million inputs,
while without block-wise optimization it ran out of memory when $p>100{,}000$.
We also notice that on a single core, 
the alternating Newton block coordinate
\begin{figure}
\centering
\includegraphics[scale=0.45]{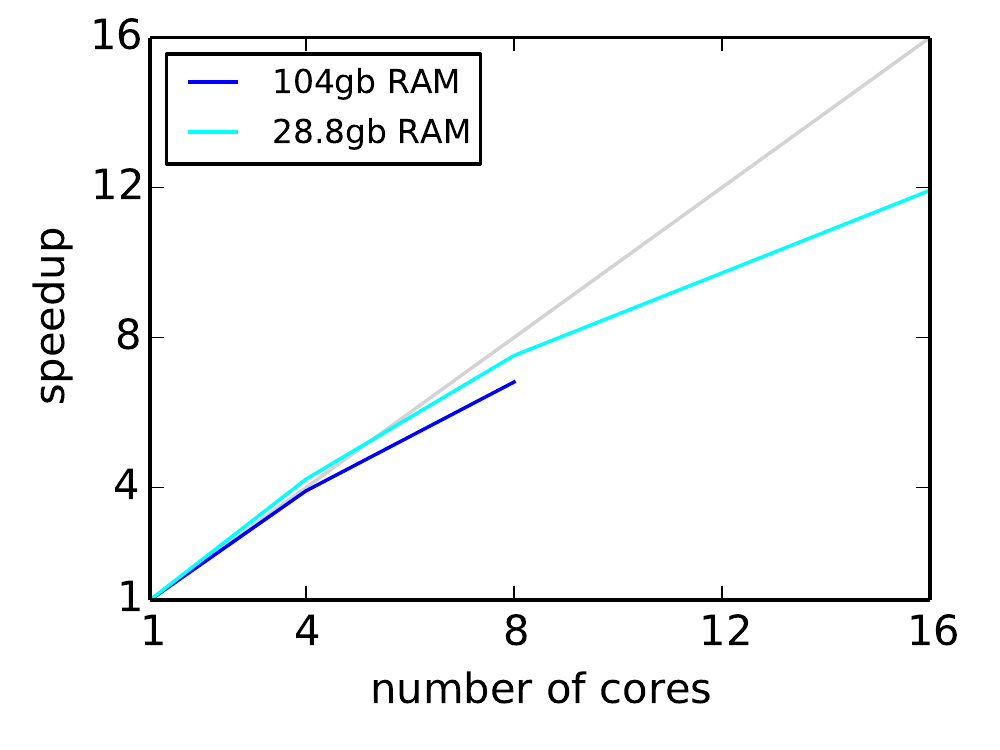}
\vspace{-9pt}
\caption{Speedup with parallelization for alternating Newton block coordinate descent.}
\vspace{-5pt}
\label{fig:parallel}
\end{figure}
descent is slighly slower 
than the same method without block-wise 
optimization because of
the need to recompute $\bSig$ and $\bSxx$. However, it is still
substantially faster than the previous method.

Finally, we evaluate the parallelization scheme for 
our alternating Newton block coordinate descent method
on multi-core machines.
Given a dataset generated from chain graph with $p=40{,}000$ and $q=20{,}000$,
in Figure \ref{fig:parallel},
we show the folds of speedup for different numbers of cores with respect to a single core. 
We obtained about 7 times speed up with 8 cores on a machine with 104Gb RAM,
and about 12 times speedup with 16 cores on a machine with 28Gb RAM.
In general, we observe greater speedup on larger problem sizes and
also for random graphs,
because such problems tend to have more cache misses that can be computed in parallel.

\subsection{Genomic Data Analysis}

We compare the different methods on a genomic dataset.
The dataset consists of genotypes for $442{,}440$ single nucleotide polymorphisms (SNPs)
and $10{,}256$ gene expression levels for 171 individuals with asthma,
after removing genes with variance under $0.01$.
We fit a sparse CGGM using SNPs as inputs
and expressions as outputs to model 
a gene network influenced by SNPs.
We also compared the methods on a smaller dataset of
$34{,}249$ SNPs from chromosome 1 and
$3{,}268$ genes with variance $>0.1$.
As typically sparse model structures are of interests
in this type of analysis, 
we chose regularization parameters 
so that the number of non-zero entries in 
each of $\bTheta$ and $\bLambda$ at convergence
was approximately $10$ times the number of genes.

The computation time of different methods are provided in Table \ref{table:real}.
On the largest problem, the previous approach 
could not run due to memory constraint, whereas our block coordinate
descent
\begin{figure}
\centering
\begin{tabular}{@{}c@{}}
\includegraphics[scale=0.47]{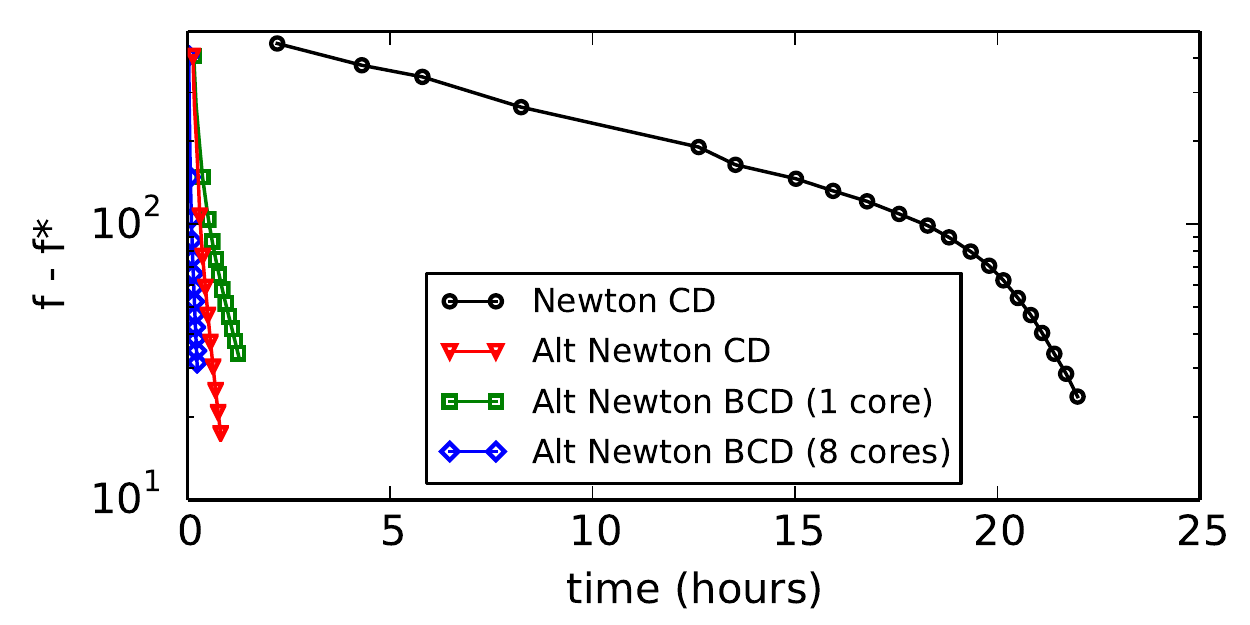} \vspace{-5pt} \\
(a) \\
\vspace{-1pt} \\
\includegraphics[scale=0.47]{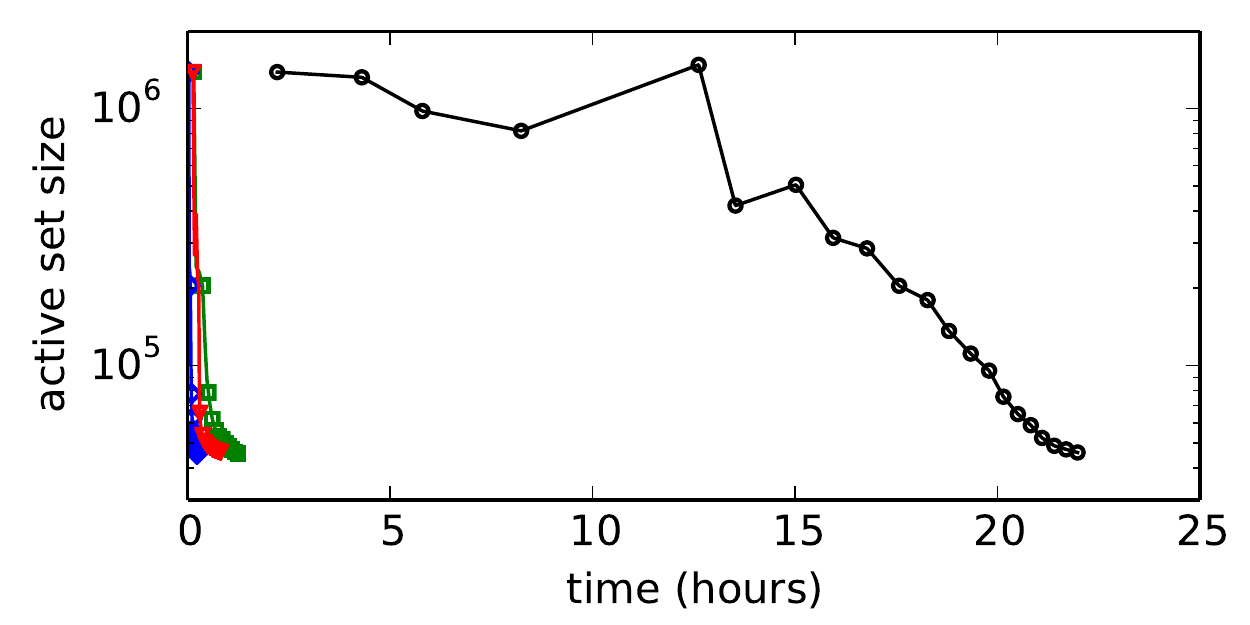} \vspace{-5pt} \\
(b) \vspace{-6pt}
\end{tabular}
\caption{Convergence results on genomic dataset.
(a) Suboptimality and (b) active set size over time.
}
\label{fig:real-convergence}
\vspace{-4pt}
\end{figure}
converged in around 11 hours.
We also compare the convergence of
the different methods on the dataset
with $34{,}249$ SNPs and $3{,}268$ gene expressions in Figure \ref{fig:real-convergence}, 
and find that our methods provide vastly superior convergence 
than the previous method.

\section{\uppercase{Conclusion}}
\label{sec:conclusion}
In this paper, we addressed the problem of large-scale optimization
for sparse CGGMs. 
We proposed a new optimization procedure, called alternating Newton 
coordinate descent, that reduces computation time
by alternately optimizing for the two sets of parameters $\bLambda$ and $\bTheta$.
Further, we extended this with block-wise optimization 
so that it can run on any machine with limited memory.

\newpage
\clearpage
\bibliography{hugecggm}{}
\bibliographystyle{abbrvnat}

\newpage
\clearpage
\onecolumn
\appendix
\section{Appendix}

\subsection{Coordinate Descent Updates for Alternating Newton Coordinate Descent Method}

In our alternating Newton coordinate descent algorithm,
each element of $\Delta_\bLambda$ is updated as follows:
\begin{align*}
(\Delta_\bLambda)_{ij} \leftarrow& (\Delta_\bLambda)_{ij} - c_\bLambda + S_{\lambda_\bLambda/a_\bLambda}(c_\bLambda-\frac{b_\bLambda}{a_\bLambda}), 
\end{align*}
where $S_{r}(w)=\textrm{sign}(w)\max(|w|-r,0)$ is the soft-thresholding operator and 
\begin{align*}
a_\bLambda =& \bSig_{ij}^2+\bSig_{ii}\bSig_{jj}+\bSig_{ii}\bPsi_{jj}+2\bSig_{ij}\bPsi_{ii} \\
b_\bLambda =& (\bSyy)_{ij}-\bSig_{ij}-\bPsi_{ij} + (\bSig\Delta_\bLambda\bSig)_{ij}
 + (\bPsi\Delta_\bLambda \bSig)_{ij} +(\bPsi\Delta_\bLambda \bSig)_{ji} \\
c_\bLambda =& \bLambda_{ij} + (\Delta_\bLambda)_{ij}.
\end{align*}
For $\bTheta$, we perform coordinate-descent updates directly on $\bTheta$ without
forming a second-order approximation of the log-likelihood to find a Newton direction,
as follows:
\begin{align*}
\bTheta_{ij} \leftarrow& \bTheta_{ij} - c_\bTheta + S_{\lambda_\bTheta/a_\bTheta}(c_\bTheta-\frac{b_\bTheta}{a_\bTheta}),
\end{align*}
where
\begin{align*}
a_\bTheta =& \bSig_{jj}(\bSxx)_{ii} \\
b_\bTheta =& 2(\bSxy)_{ij} + 2(\bSxx\bTheta\bSig)_{ij} \\
c_\bTheta =& \bTheta_{ij}.
\end{align*}

\begin{figure*}[h!]
\centering
\begin{tabular}{@{}c@{\hspace{10pt}}c}
\includegraphics[scale=0.45]{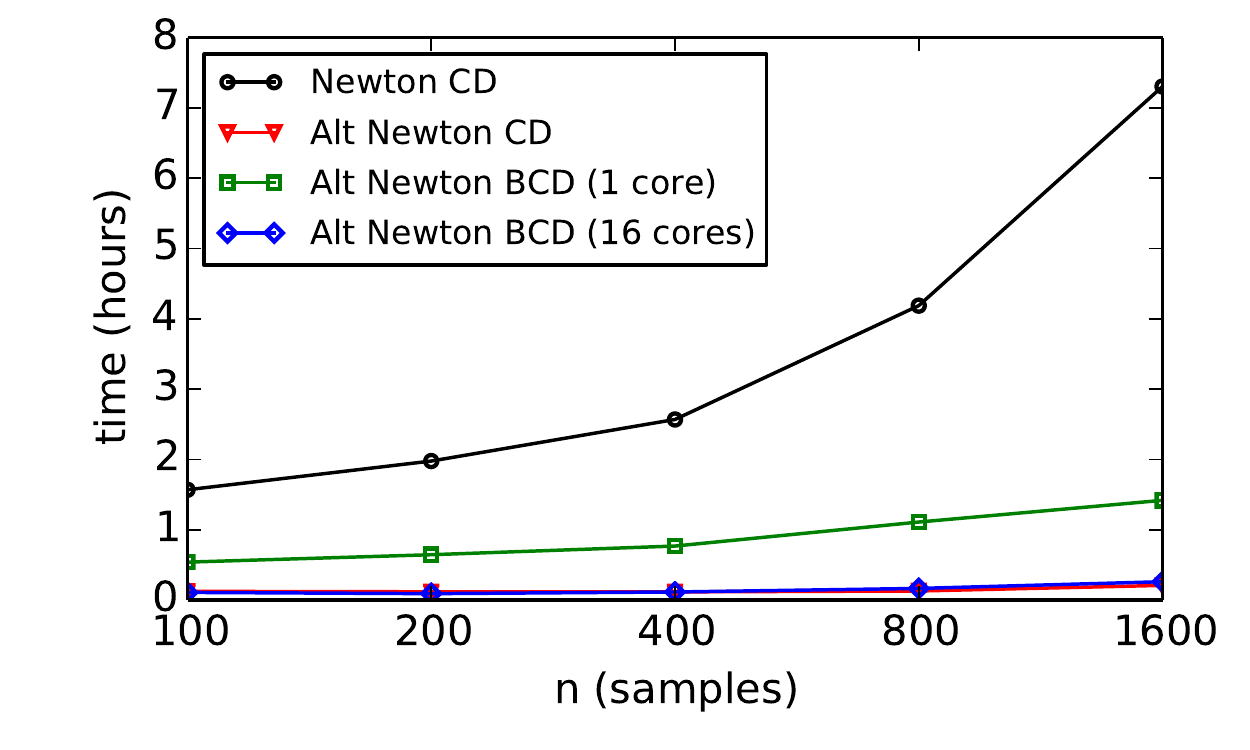}
& \includegraphics[scale=0.45]{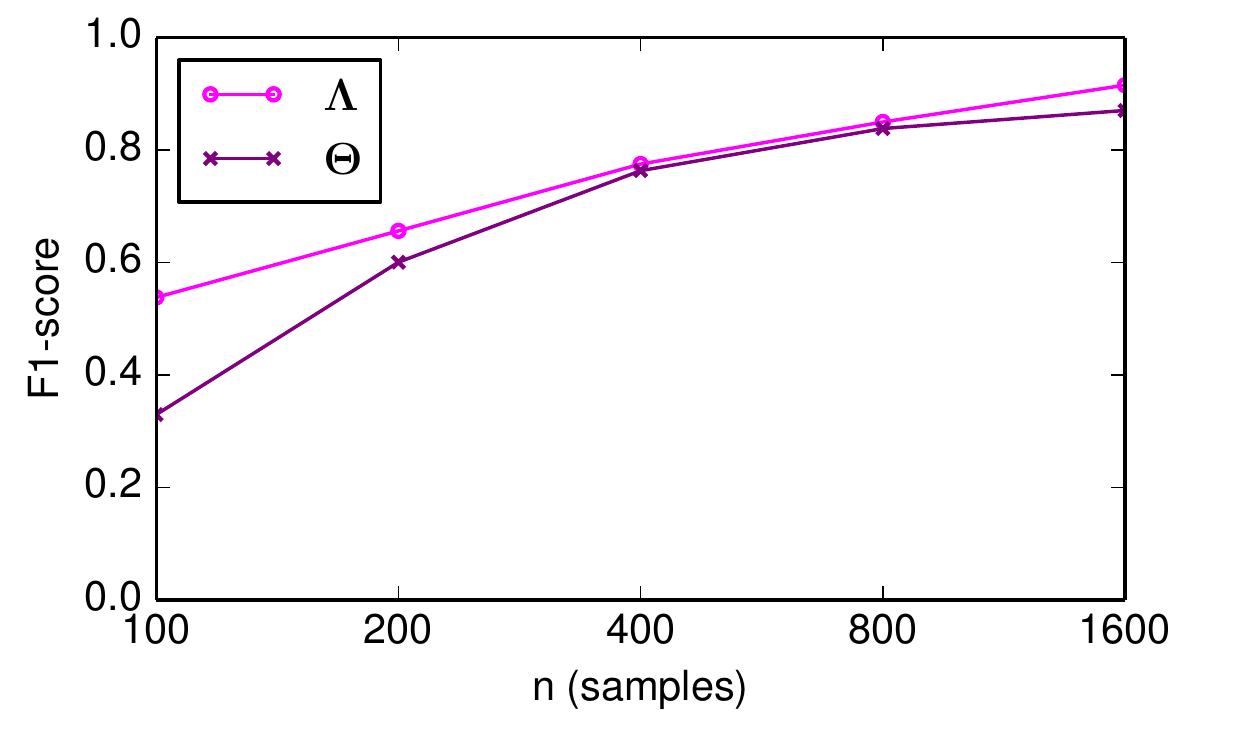} \\
\vspace{-3pt}
(a) & (b)
\end{tabular}
\vspace{-0.05in}
\caption{Results from varying sample size $n$ on chain graph with $p=q=10{,}000$.
(a) Comparison of computation time of different methods.
(b) Comparison of edge recovery accuracy as measured by $F_1$-score. 
}
\label{fig:scaling-n}
\end{figure*}

\subsection{Additional Results from Synthetic Data Experiments}

We compare the performance of the different algorithms on synthetic datasets
with different sample sizes $n$, using a chain graph structure 
with $p=q=10{,}000$.
Figure \ref{fig:scaling-n}(a) shows that our methods run significantly faster
than the previous method across all sample sizes. 
In Figure \ref{fig:scaling-n}(b) we measure the accuracy in recovering
the true chain graph structure in terms of $F_1$-score for different sample sizes $n$.
At convergence, $F_1$-score was the same for all methods to three significant digits.
As expected, the accuracy improves as the sample size increases.

\end{document}